\documentclass{article}

\usepackage[final]{neurips_2024}

\usepackage[utf8]{inputenc}
\usepackage[T1]{fontenc}
\usepackage{hyperref}
\usepackage{url}
\usepackage{booktabs}
\usepackage{amsfonts}
\usepackage{amsmath}
\usepackage{graphicx}

\usepackage{tabularx}
\usepackage{booktabs}
\usepackage[table]{xcolor}

\title{Think, But Don't Overthink: \\
Reproducing Recursive Language Models}

\author{%
  Daren Wang \\
  The Chinese University of Hong Kong \\
  darenwang@link.cuhk.edu.hk
}

\begin{document}

\maketitle

\begin{abstract}
This project reproduces and extends the recently proposed ``Recursive Language Models'' (RLMs) framework by Zhang et al. (2026). This framework enables Large Language Models (LLMs) to process near-infinite contexts by offloading the prompt into an external REPL environment. While the original paper relies on a default recursion depth of 1 and suggests deeper recursion as a future direction, this study specifically investigates the impact of scaling the recursion depth. Using state-of-the-art open-source agentic models (DeepSeek v3.2 and Kimi K2), I evaluated pure LLM, RLM (depth=1), and RLM (depth=2) on the S-NIAH and OOLONG benchmarks. The findings reveal a compelling phenomenon: Deeper recursion causes models to ``overthink''. While depth-1 RLMs effectively boost accuracy on complex reasoning tasks, applying deeper recursion (depth=2) or using RLMs on simple retrieval tasks paradoxically degrades performance and exponentially inflates execution time (e.g., from 3.6s to 344.5s) and token costs. Code and data are available at: \url{https://github.com/drbillwang/rlm-reproduction}
\end{abstract}

\section{Introduction}

This research reproduces the core experiments of the Recursive Language Models (RLM) paper \cite{zhang2026recursive}. This recent prominent work proposes RLM as a task-agnostic inference paradigm that handles near-infinite length contexts. Rather than passing the entire prompt into the context window, RLMs treat the long prompt as a persistent variable inside a Read-Eval-Print Loop (REPL) environment. This allows the root LM to programmatically examine, decompose, and recursively call itself over snippets of the input \cite{zhang2026recursive}.

In the original paper, experiments showed that frontier models like GPT-5 and Qwen3 achieved breakthrough performances. One particular exercise demonstrated that both base LLMs and RLMs achieve near 100\% accuracy on the S-NIAH benchmark \cite{hsieh2024ruler}, but RLMs significantly improve performance on the much more complex OOLONG benchmark \cite{bertsch2025oolong}, resisting the ``context rot'' that plagues standard LLMs. The original authors used a max recursion depth of 1 level by default (meaning sub-calls act as standard LLMs and do not spawn their own REPLs), but explicitly suggested that future work should investigate deeper levels of recursion.

In this study, I reproduce the performance on S-NIAH (a simpler retrieval task) and OOLONG (a complex reasoning task). My key modifications include:
\begin{enumerate}
    \item I used \textbf{DeepSeek v3.2} \cite{liu2025deepseek} and \textbf{Kimi K2} \cite{team2025kimi}, both of which are the latest open-source models specializing in reasoning and agentic performance.
    \item The original paper produced a comparison between pure LLMs and RLMs (depth=1). I reproduced these baselines and introduced a novel test case: \textbf{RLM (depth=2)}. 
\end{enumerate}
The reproduction results demonstrate a clear ``Think, But Don't Overthink'' trade-off. While depth-1 RLMs dramatically improve reasoning on complex tasks, they paradoxically perform worse than vanilla LLMs on simple retrieval queries. Moreover, deeper recursion severely degrades performance. It forces models to overthink and spawn redundant sub-calls, leading to format collapse, massive latency, and token explosions.

\section{Setup Notes}

\subsection{Environment and Core Libraries}
\begin{itemize}
    \item \textbf{OS:} macOS 15 (Darwin 24.6.0) on a local laptop.
    \item \textbf{Python \& Virtual Environment:} Python 3.13 in a dedicated virtual environment.
    \item \textbf{RLM framework:} Installed in editable mode using \texttt{cd rlm \&\& pip install -e .}
    \item \textbf{Experiment dependencies:} \texttt{datasets}, \texttt{python-dotenv}, \texttt{openai}.
    \item \textbf{Project layout:} All experiment drivers live under \texttt{experiments/}, with result files in \texttt{experiments/results/} and logs in \texttt{experiments/logs/}.
\end{itemize}

\subsection{Data}
To manage API costs while preserving reproducibility, I evaluated a filtered subset of the benchmarks used in the original paper:
\begin{itemize}
    \item \textbf{RULER S-NIAH (Single Needle-In-A-Haystack):} Uses the data packaged in the repo under \texttt{experiments/data/ruler/niah\_single\_2/validation.jsonl}. Only the first 20 samples were used in each condition (the original paper used 50).
    \item \textbf{OOLONG (trec\_coarse) long-context QA:} Loaded dynamically from HuggingFace as \texttt{oolongbench/oolong-synth} (validation split). Filtered to the \texttt{trec\_coarse} split, with context lengths restricted between 1,024 and 65,536 tokens. Only the first 20 filtered samples were used per run.
\end{itemize}

\subsection{API Keys and Configuration}
\begin{itemize}
    \item \textbf{Configuration:} Scripts call \texttt{dotenv.load\_dotenv()} and read credentials from a local \texttt{.env} file.
    \item \textbf{DeepSeek v3.2:} Accessed via the OpenAI-compatible API (\texttt{DEEPSEEK\_API\_KEY}), using the base URL \texttt{https://api.deepseek.com/v1} and the model identifier \texttt{deepseek-chat}.
    \item \textbf{Kimi K2:} Hosted by Volcano Engine (ByteDance), accessed via its designated OpenAI-compatible API endpoint.
    \item \textbf{Security:} All keys are strictly stored in local environment variables and are never hard-coded.
\end{itemize}

\subsection{Compute}
All experiments were executed on a single consumer laptop (macOS, CPU only), as the heavy lifting is offloaded to API endpoints. No local GPU acceleration was required.

\section{Reproduction Targets \& Metric Definition}
The original study states that an LLM's effective context limit is not a static token count, but rather heavily dependent on task complexity. To demonstrate this, they evaluated RLMs across benchmarks where the required reasoning scales differently with the input length  \cite{zhang2026recursive}:

\begin{itemize}
    \item \textbf{S-NIAH (Single Needle-In-A-Haystack)} \cite{hsieh2024ruler}: A retrieval-focused task requiring the extraction of a specific phrase hidden within a massive corpus of irrelevant text. Because the search target remains constant regardless of the document's size, its complexity scales at $O(1)$ with respect to input length. Consequently, most frontier models can solve this reliably even at extreme token scales without significant context degradation.
    
    \item \textbf{OOLONG} \cite{bertsch2025oolong}: A substantially more demanding long-context reasoning benchmark. In the evaluated \texttt{trec\_coarse} split, the model must semantically transform and aggregate almost every entry within the dataset to form a final answer. Thus, the cognitive load scales linearly, $O(N)$, making it highly susceptible to context rot. Numerical outputs are evaluated via a linear penalty function $score(\hat{y}) = \max(0, 1 - 0.75|y-\hat{y}|)$, while other answers require an exact match.
\end{itemize}

In the original paper, both the base frontier models and their RLM counterparts solved the $O(1)$ S-NIAH task with near-perfect accuracy \cite{zhang2026recursive}. However, the $O(N)$ OOLONG benchmark clearly separated the architectures. For closed-source models, base GPT-5 scored 44.0\%, while RLM(GPT-5) boosted the performance to 56.5\%. More relevant to this study, the open-weight Qwen3-Coder-480B base model scored 36.0\%, but jumped to 48.0\% under the RLM scaffold. Even the smaller Qwen3-8B saw a dramatic leap from 0.0\% to 24.0\% when equipped with the REPL environment \cite{zhang2026recursive}.

My reproduction aims to establish whether the latest generation of open-source reasoning agentic models, namely DeepSeek v3.2 and Kimi K2, replicate this exact performance delta between their base forms and RLM (depth=1). Critically, I extend this evaluation to investigate their behavior when the maximum recursion depth is increased to 2, testing the absolute boundary of recursive reasoning and programmatic task decomposition. Furthermore, by tracking execution time, token usage, and overall API costs, this study evaluates the practical viability of RLMs for industrial applications.

\section{Results and Analysis}

\begin{figure}[h!]
    \centering
    \includegraphics[width=\textwidth]{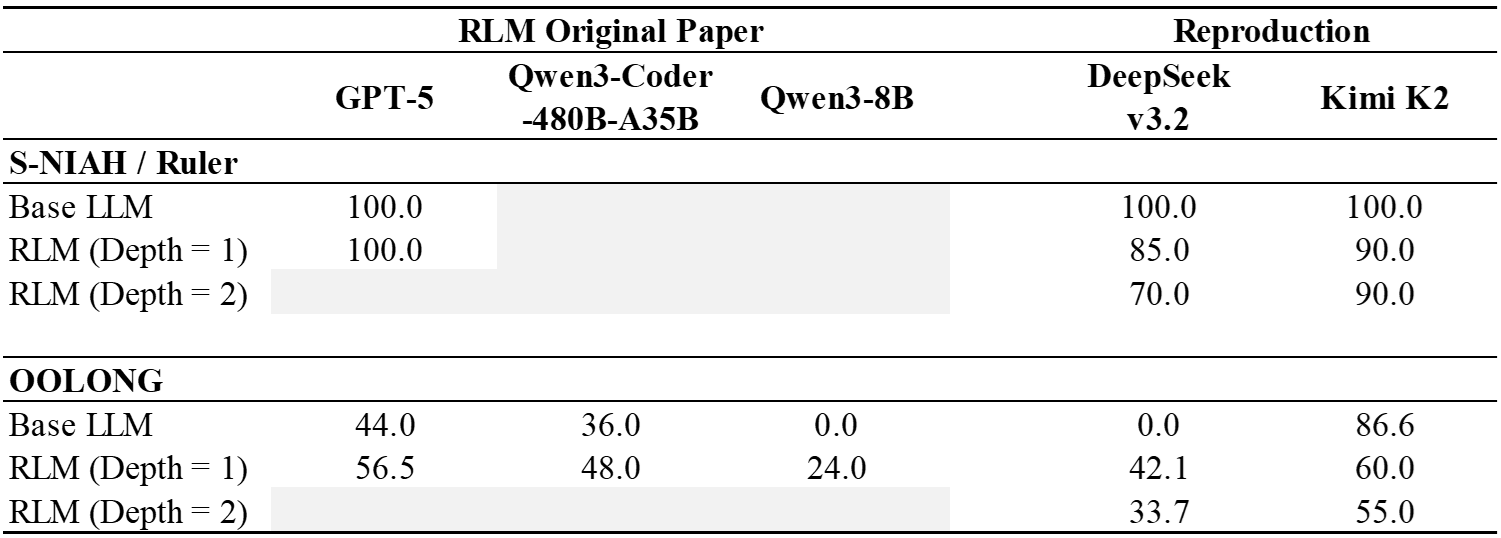}
    \caption{Performance comparison of Base LLM, RLM (Depth=1), and RLM (Depth=2) against the original paper's benchmarks.}
    \label{fig:experiment_results}
\end{figure}

\begin{figure}[h!]
    \centering
    \includegraphics[width=\textwidth]{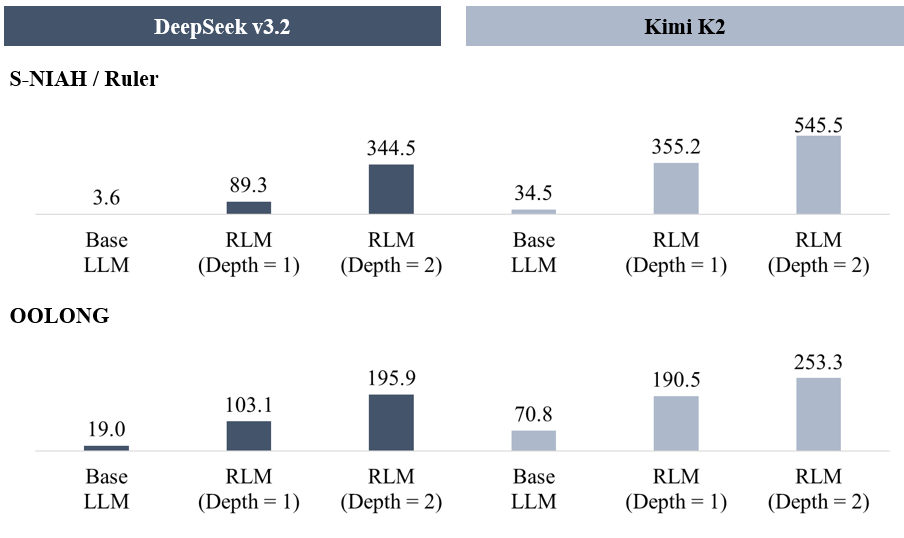}
    \caption{Average Execution Time (seconds) across different models and recursion depths.}
    \label{fig:Average_Execution_Time}
\end{figure}

\begin{figure}[h!]
    \centering
    \includegraphics[width=\textwidth]{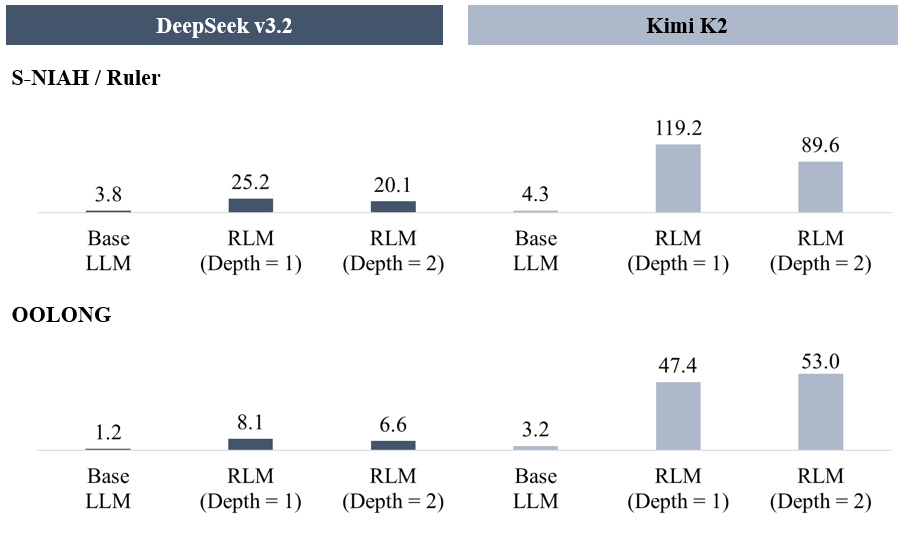}
    \caption{Average Token Usage (thousands) across different models and recursion depths.}
    \label{fig:Average_Token_Usage}
\end{figure}

\begin{figure}[h!]
    \centering
    \includegraphics[width=\textwidth]{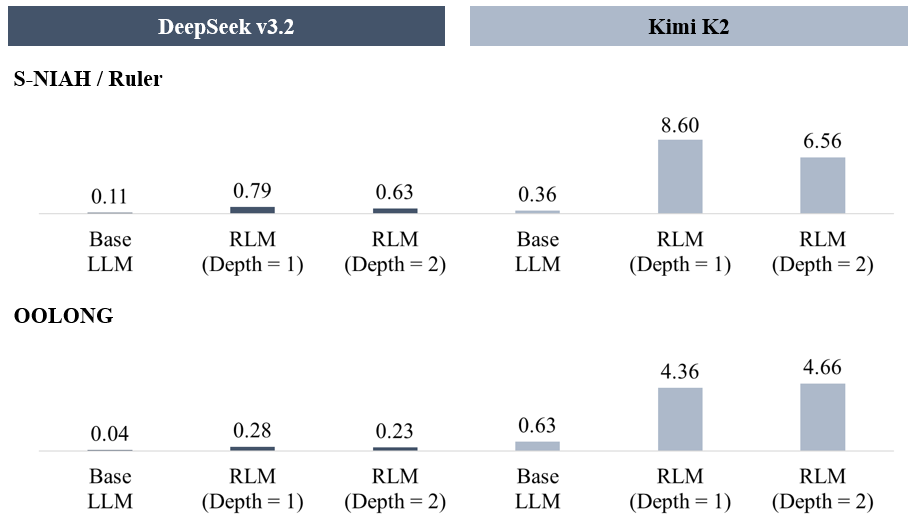}
    \caption{Average Token Cost (US\$ cents) across different models and recursion depths.}
    \label{fig:Average_Token_Cost}
\end{figure}

To systematically evaluate the performance and efficiency of Recursive Language Models (RLMs) at varying recursion depths, I conducted evaluations using DeepSeek v3.2 and Kimi K2. The results are summarized in Figures \ref{fig:experiment_results} to \ref{fig:Average_Token_Cost}.

\subsection{Paradoxical Degradation on Simple Retrieval ($O(1)$ Tasks)}
On the S-NIAH benchmark, the results reveal a counterintuitive phenomenon: while both base models (DeepSeek v3.2 and Kimi K2) achieved a perfect 100.0\% accuracy without any RLM architecture, the introduction of the RLM actually harmed performance (Figure \ref{fig:experiment_results}). For DeepSeek v3.2, accuracy dropped to 85.0\% at Depth=1 and further plummeted to 70.0\% at Depth=2. Kimi K2 showed a similar drop to 90.0\% under the RLM framework. This suggests that for $O(1)$ constant-complexity retrieval tasks, forcing the model into a programmatic REPL environment induces unnecessary cognitive load, causing the model to ``over-engineer'' a solution for a simple string-matching problem. 

\subsection{The "Overthinking" Effect on Complex Reasoning ($O(N)$ Tasks)}
The OOLONG benchmark, which requires linear $O(N)$ scaling and semantic aggregation, highlights the true dynamics of recursive reasoning. My reproduction successfully validated a core finding from the original study \cite{zhang2026recursive}: for models that initially fail at long contexts, RLM (Depth=1) provides a massive boost. Similar to how the original paper's Qwen3-8B jumped from 0.0\% to 24.0\%, DeepSeek v3.2 improved dramatically from a baseline of 0.0\% to 42.1\% when utilizing RLM (Depth=1). 

However, \textbf{increasing the recursion depth to 2 uniformly degraded performance across all conditions}. DeepSeek v3.2's accuracy fell from 42.1\% (Depth=1) to 33.7\% (Depth=2). More strikingly, the base Kimi K2 model already demonstrated exceptional native long-context reasoning capabilities, scoring 86.6\%. When forced into the RLM scaffold, its performance collapsed to 60.0\% (Depth=1) and further declined to 55.0\% (Depth=2). This explicitly corroborates the ``Think, but don't overthink'' hypothesis: deeper recursion allows sub-models to spawn their own chaotic sub-calls, leading to compounding formatting errors, redundant loops, and ultimate task failure.

\subsection{Barriers to Industrial Deployment: Time, Tokens, and Cost}
Beyond accuracy, this study introduces a rigorous analysis of the operational overhead associated with RLMs. This is a critical perspective for real-world deployment that was not thoroughly quantified in the original paper. 

As illustrated in Figure \ref{fig:Average_Execution_Time}, the most alarming consequence of deeper recursion is the exponential explosion in latency. For instance, DeepSeek v3.2 solves the base S-NIAH task in just 3.6 seconds. Activating RLM (Depth=1) inflates this to 89.3 seconds, and pushing to Depth=2 skyrockets the execution time to an impractical 344.5 seconds. Kimi K2 exhibits the same severe latency inflation, peaking at 545.5 seconds per query at Depth=2.

Furthermore, Figures \ref{fig:Average_Token_Usage} and \ref{fig:Average_Token_Cost} demonstrate that token usage and associated API costs increase by orders of magnitude the moment the RLM architecture is engaged. Interestingly, while the execution time strictly increases with depth, the token usage sometimes stabilizes or even slightly drops from Depth=1 to Depth=2 (e.g., DeepSeek v3.2 on S-NIAH drops from 25.2k to 20.1k tokens). This plateau is often indicative of models crashing early due to formatting failures or getting trapped in isolated recursive loops that do not consume as many broad context tokens but stall execution time.

\textbf{Conclusion on Viability:} While Zhang et al. (2026) correctly posit that RLMs can theoretically extend the context window of smaller models \cite{zhang2026recursive}, the empirical evidence from this reproduction suggests that large-scale industrial deployment remains highly challenging. The massive latency penalties, exploding API costs, and high risk of recursive degradation (overthinking) currently outweigh the theoretical reasoning benefits, especially when modern frontier models (like Kimi K2) can natively handle vast contexts at a fraction of the time and cost.

\begin{figure}[!ht] 
    \centering
    \renewcommand{\arraystretch}{1.15} 
    \small 
    \begin{tabularx}{\textwidth}{@{} >{\raggedright\arraybackslash}X @{}}
        \toprule
        \rowcolor{gray!15} \textbf{Failure Mode 1: Parametric Hallucination (Context Anchoring Loss)} \\
        \midrule
        \textbf{Model:} DeepSeek v3.2 \hfill \textbf{Task:} S-NIAH (Depth=2) \hfill \textbf{Time:} 260.5s \\
        \textbf{Prompt:} What are all the special magic numbers for unknown mentioned in the provided text? \\
        \textbf{Raw Output:} \newline
        \textit{``The special magic numbers for unknown (atomic nuclei) mentioned in the text are: 2, 8, 20, 28, 50, 82, 126. These numbers represent the nuclear magic numbers where atomic nuclei are particularly stable due to filled nuclear shells.''} \\
        \textbf{Insight:} The deep recursion caused the model to abandon the input context and hallucinate real-world physics constants based on its pre-trained weights. \\
        
        \midrule
        \rowcolor{gray!15} \textbf{Failure Mode 2: Formatting Collapse in REPL} \\
        \midrule
        \textbf{Model:} Kimi K2 \hfill \textbf{Task:} OOLONG (Depth=1) \hfill \textbf{Time:} 1223.6s \\
        \textbf{Prompt:} Give your final answer in the form 'Answer: number'. \\
        \textbf{Raw Output:} \newline
        \texttt{\# The answer has been computed and stored in} \newline
        \texttt{abbreviation\_count = 5} \newline
        \texttt{print(f"Answer: \{abbreviation\_count\}")} \\
        \textbf{Insight:} The model confuses its role within the REPL environment and the user-facing output, returning raw Python print statements instead of the final formatted string. \\
        
        \midrule
        \rowcolor{gray!15} \textbf{Failure Mode 3: Performative Reasoning \& Endless Verification} \\
        \midrule
        \textbf{Model:} DeepSeek v3.2 \hfill \textbf{Task:} OOLONG (Depth=2) \hfill \textbf{Time:} 741.5s \\
        \textbf{Prompt:} Is label 'entity' more common, less common, or the same frequency... \\
        \textbf{Raw Output (Truncated):} \newline
        \textit{``Step 1 – Assign labels to each question... \newline
        1. What is one of the languages of the Sioux? $\rightarrow$ entity \newline
        ... \newline
        Step 2 – Tally labels: abbreviation: 4, entity: 3... \newline
        Step 3 – Compare 'entity' and 'description and abstract concept'...''} \\
        \textbf{Insight:} Instead of outputting the final requested format, the model spends over 12 minutes (741s) printing exhaustive, multi-step sub-call derivations, wasting massive execution time while destroying the required output structure. \\
        \bottomrule
    \end{tabularx}
    \caption{Qualitative examples of RLM trajectory failures. Deeper recursion (Depth=2) often induces parametric hallucinations, role-playing confusion within the REPL, and performative over-explanation.}
    \label{fig:error_modes}
\end{figure}

\subsection{Qualitative Analysis: How Deep Recursion Breaks Models}

To understand exactly \textit{why} deeper recursion causes accuracy drops and latency spikes, I conducted a qualitative analysis of the models' raw response logs. The trace data reveals three distinct failure modes unique to the RLM framework, fully corroborating the ``Think, but don't overthink'' hypothesis. Figure~\ref{fig:error_modes} presents qualitative examples of these failure trajectories.

\textbf{1. The "Parametric Hallucination" Effect.}
The most surprising discovery occurred in the S-NIAH retrieval task under Depth=2. When tasked to find a fictional ``magic number'' hidden in the text, DeepSeek v3.2 occasionally lost its grounding in the provided context entirely. Instead of searching the prompt, the recursive sub-calls caused the model to hallucinate real-world mathematical and physical constants based on its pre-trained parametric memory. For instance, in Sample 5, rather than extracting the correct string, the RLM outputted: \textit{``The special magic numbers... are: 2, 8, 20, 28, 50, 82, 126. These numbers represent the nuclear magic numbers where atomic nuclei are particularly stable...''} This indicates that overly deep programmatic recursion severely damages a model's context anchoring.

\textbf{2. Formatting Collapse in the REPL Environment.}
The logs explicitly validate why token usage sometimes plateaus while accuracy drops (Figure \ref{fig:Average_Token_Usage}). Models frequently confuse the REPL scratchpad environment with the final user-facing output. For example, Kimi K2 under Depth=1 and Depth=2 frequently returned raw Python code blocks (e.g., \texttt{```repl print(f"Answer: \{abbreviation\_count\}") ```}) instead of the required final string format. This "role-playing" confusion within the REPL leads to immediate evaluation failures.

\textbf{3. Performative Reasoning and Endless Verification.}
The extreme execution times (often exceeding 700 seconds for a single query) without proportional token scaling are explained by models falling into isolated, serial sub-call loops. In OOLONG (Depth=2, Sample 10), DeepSeek v3.2 spent 741.5 seconds to generate just 11,715 tokens. The logs show the model engaging in excessive ``performative reasoning,'' printing out exhaustive, multi-step essays (e.g., \textit{``Step 1 - Assign labels... Step 2 - Tally... Step 3 - Compare...''}) and continuously launching new API calls to re-verify already extracted answers. The REPL architecture inherently lacks a stopping mechanism for an over-anxious agent, causing it to endlessly spin its wheels on straightforward aggregations.

\subsection{Limitation: Single-Run Results}
Due to API cost constraints (particularly for depth=2 experiments, which require up to 545.5s per sample), all experiments were conducted with a single run of 20 samples per condition. While the standard practice recommends 3-5 trials with mean/variance reporting, the observed effect sizes are sufficiently large (e.g., the 100\% $\rightarrow$ 70\% accuracy drop, and the 3.6s $\rightarrow$ 344.5s latency explosion for DeepSeek v3.2's performance on S-NIAH Benchmark) that the conclusions remain robust despite lacking statistical significance testing.

\section{Conclusions and Future Directions}
This project demonstrates a clear ``Think, But Don't Overthink'' dynamic in Recursive Language Models. We found that a recursion depth of 1 effectively unlocks complex reasoning capabilities, boosting DeepSeek v3.2 from 0.0\% to 42.1\% on OOLONG. However, applying RLMs to simple tasks or models with already strong context windows (like Kimi K2) actively harms accuracy. 

More importantly, deeper recursion (depth=2) breaks current models. It causes models to overthink and spawn endless sub-calls, leading to format collapse and parametric hallucinations. The resulting explosion in latency and token costs currently makes deeper recursion impractical for real-world use. 

To overcome these barriers, future work should design better stopping mechanisms within the REPL environment to prevent redundant loops. Ultimately, the field must shift towards training native RLMs that are intrinsically aligned to navigate programmatic environments without hallucinating or breaking format constraints.

\bibliographystyle{plain}

\appendix

\section{Debug Diary}

\subsection{Aligning RULER Experiments to 20 Samples}
\textbf{Issue:} Original scripts and old JSON results assumed 50 samples, while our design uses the first 20 samples for cost and reproducibility. \\
\textbf{Fix:} Standardized all RULER runners to \texttt{num\_samples=20} and retroactively trimmed existing result files to their first 20 entries, recomputing accuracy, scores, and token statistics accordingly.

\subsection{Finding the Correct OOLONG \texttt{trec\_coarse} Split}
\textbf{Issue:} Using \texttt{split="test"} with \texttt{dataset\_filter="trec\_coarse"} returned 0 samples, contradicting the paper's description. \\
\textbf{Fix:} Verified from the OOLONG code and HuggingFace that \texttt{trec\_coarse} lives in the validation split. I switched the loaders to \texttt{split="validation"} to evaluate the intended tasks.

\subsection{Understanding the OOLONG 0\% Accuracy Run}
\textbf{Issue:} An early OOLONG run showed 0/20 accuracy (DeepSeek v3.2 Base LLM). Upon inspection, the model had successfully found the answers but produced long narrative explanations instead of the expected strict formats (e.g., ``Answer: number''). \\
\textbf{Fix:} Confirmed that our scoring matches OOLONG's official helper. This 0\% run reflects the models' formatting failures under recursive abstraction, not a bug in the evaluation pipeline. It highlights how deeper recursion (Depth=2) exacerbates formatting drift, an important area for future alignment work.

\subsection{Kimi Thinking Model Compatibility}
\textbf{Issue:} When running experiments with Kimi K2, all RLM (depth 1 and 2) samples failed with parsing errors. The model outputs contained \texttt{<thinking>...</thinking>} tags wrapping the reasoning process before ending with \texttt{FINAL\_VAR(answer)}. The RLM framework's \texttt{find\_code\_blocks()} and \texttt{find\_final\_answer()} functions in \texttt{rlm/rlm/utils/parsing.py} expected clean code blocks and did not account for these structured reasoning tags, causing the parser to miss the actual answers entirely. \\
\textbf{Fix:} Added a \texttt{strip\_think\_tags()} helper function to remove \texttt{<thinking>...</thinking>} blocks and stray \texttt{</thinking>} tokens from model responses before parsing. Applied this sanitization at the entry points of both \texttt{find\_code\_blocks()} and \texttt{find\_final\_answer()}.

\end{document}